# Satellite Capture and Servicing Using Networks of Tethered Robots Supported by Ground Surveillance

**Himangshu Kalita**
*Space and Terrestrial Robotic Exploration Laboratory,
Department of Aerospace and Mechanical Engineering, University of Arizona*
**Roberto Furfaro**
*Department of Systems and Industrial Engineering, University of Arizona*
**Jekan Thangavelautham**
*Space and Terrestrial Robotic Exploration Laboratory,
Department of Aerospace and Mechanical Engineering, University of Arizona*

## ABSTRACT

There is ever growing demand for satellite constellations that perform global positioning, remote sensing, earth-imaging and relay communication. In these highly prized orbits, there are many obsolete and abandoned satellites and components strewn posing ever-growing logistical challenges. This increased demand for satellite constellations pose challenges for space traffic management, where there is growing need to identify the risks probabilities and if possible mitigate them. These abandoned satellites and space debris maybe economically valuable orbital real-estate and resources that can be reused, repaired or upgraded for future use. On-orbit capture and servicing of a satellite requires satellite rendezvous, docking and repair, removal and replacement of components. Launching a big spacecraft that perform satellites servicing is one credible approach for servicing and maintaining next-generation constellations.

By accessing abandoned satellites and space debris, there is an inherent heightened risk of damage to a servicing spacecraft. Under these scenarios, sending multiple, small-robots with each robot specialized in a specific task is a credible alternative, as the system is simple and cost-effective and where loss of one or more of robot does not end the mission. Inherent to this network of small robots is the need for ground surveillance and observation of the system both to provide real-time information about the space debris, in addition to providing position, navigation and tracking support capabilities. Eliminating the need for a large spacecraft or positioning the large spacecraft at safe distance to provide position, navigation and tracking support simplifies the system and enable the approach to be extensible with the latest ground-based sensing technology. In this work, we analyze the feasibility of sending multiple, decentralized robots that can work cooperatively to perform capture of the target satellite as first steps to on-orbit satellite servicing.

We further analyze the extent of how a ground-based civilian surveillance system can be used to provide real-time observation support in place of using a larger, on-orbit servicing mothership. The multi-robot system will be deployed in a formation interlinked with spring-tethers in one of several configurations include an 'x' configuration. These tethered small robots will perform one-time autonomous rendezvous, capture and servicing of satellites in LEO and GEO orbits. Use of spring tether enables dynamic capture of a target object that maybe freely tumbling or travelling at different velocities in the range of 15 m/s or more. The tether enables converting a translational motion into an angular snagging motion much like bola used by prehistoric hunters to snag a prey. Using multiple tethered robots, it may be possible to apply differential control to capture a spacecraft under more desirable angular and linear velocities. However, there also exists challenges of mitigating tangled tethers. The option exists for each robot to disconnect from its tether to avoid complex tangled scenarios.

The tethers may be rolled up to shorten or lengthen the cord length between these robots. After docking with the target satellite, each robot secures itself on the satellites surface using spiny gripping actuators. The multi-robot system can crawl on the satellite's surface with each robot moving one by one using rolling and hopping mobility capabilities. If any robot loses grip, the multirobot system with robots anchored to the surface keeps the entire system secure. The system can also be used to carry larger components and place them on a specific location on a target satellite. The rolling up of the tethers enables fine level position control of a larger object. Through this distributed controls approach, the risk is distributed, and a collective of small robots can perform multiple servicing tasks on the satellite simultaneously. A variation of this scenario is to use these small-robots to perform assembly of large passive space structures and warehousing of large space structures.

## 1. INTRODUCTION

The growing number of satellite missions to Low Earth Orbit (LEO) is evidence of a maturing industry, where a multitude of new applications and commercial needs are being fulfilled. However, this rapidly growing industry also imposes challenges, namely congestion, increase space debris and the ever increasing need for space traffic management. The growing number of space missions and the space junk left behind if left unchecked can have catastrophic consequences resulting in cascading impact of space debris, known as the "Kessler Effect" [14] that ultimately cuts off access to LEO and becomes a barrier to space.

Use of specialized space robots to rendezvous, capture and dock with satellites to service them is viable for satellites that were designed for this capability. The vast majority of spacecraft were not designed to be captured or serviced in this manner and hence the rendezvous and capture maneuvers pose risks. Tether technology enables for capture, de-orbit and boosting of satellites in orbit. The advantage of a tether is that the target satellite can be kept at a distance at first from the main valuable servicing satellite before performing preparatory service, repair or deconstruction work. With the target satellite being at distance, it could be carefully surveyed to determine whether components or substances are spilling or prone to dispersion upon contact. The tether maybe used to bring the two satellites where one maybe derelict to the same orbit, relative velocity and attitude.

Thus tethering target satellite with the service satellite shows some important advantage to satellite servicing, orbit debris capture and clean-up. However, the remaining challenge is how to affix the tether to a target satellite that may have never been designed for servicing, that maybe already be damaged or is disintegrating due to prolonged corrosive forces (due to atomic oxygen) in Earth orbit. In this paper, we proposed sending small teams of microbots affixed with tethers, forming an 'x' configuration to rendezvous and then land and crawl onto the surface of the target satellite. This paper is an update to our earlier concept paper on this topic [1]. In this paper, we extensively look at the potential applications of the tethering approach and how it is advantageous to satellite capture and servicing in addition to the post-capture surveying task. The microbots would then crawl over the satellite surface to find suitable locations to affix a tether to the servicing satellite. Our nominal scenario consists of four microbots in an 'x' configuration to first propel towards the target satellite and then land on to its surface to then prepare attaching an external tether. Using these four microbots, it is possible to effectively wrap tethers around the entire target satellite and avoid just having to use one contact point that upon increased tensile force from the service satellite may break off.

Having 4 or more microbots perform this critical task allows for risk to be transferred to the small, redundant, dispensable units that can be readily be replaced. The microbots with their low mass of 3 kg or less also minimize risk of impact with the target satellite and resultant damage. Furthermore, the microbots may have crucial role to actively survey the target satellite prior to or after capture. This staged concept of deploying microbots to inspect, rendezvous and wrap around tethers for a large service robot to then proceed with its main task overall minimizes overall mission risk, and enables capture, servicing of a wider range of satellites that may not have readily been designed for rendezvous and docking for servicing. In the following sections, we present background to tethering and potential applications, tether dynamics modelling, capture and docking simulation results, followed by post-capture maneuvers and surveying, conclusions and future work.

## 2. BACKGROUND

Tethering of a spacecraft is not a new concept and has been present since the 1960's with Gemini VI and VII tether experiments [3, 8]. This was followed by more in-depth experiments during the space shuttle era in the 1980s and 1990s. This included Charge-1 in 1983, Charge-2 in 1985, OEDIPUS-A (Observations of Electric-Field Distribution in the Ionospheric Plasma—a Unique Strategy) in 1989 [3] and OEDIPUS-C in 1995 [3], TSS-1 in 1992 and TSS-1R in 1996 [3], SEDS-1 (Small Expendable Deployer System) in 1993 and SEDS-2 in 1994 [3], PMG (Plasma Motor Generator) in 1993 [3], TiPS (Tether Physics and Survivability) experiments in 1996 [3], YES (Young Engineers Satellite) in 1997 [6], YES2 in 2007 [7], ATEx (Advanced Tether Experiment) in 1998 [9], MAST (Multi-Application Survivable Tether) in 2007 [10].

These experiments tested tether deployment, attitude control stabilization, gravity-gradient stabilization, power generation, drag-generation for de-orbiting and boosting/reboosting to higher orbits [3-4,8]. Tethered experiments have also been used to generate artificial gravity, facilitate payload rendezvous and capture and have been shown to enable aero-assisted maneuvering. The versatility of using tethers for a wide variety of applications make them an important technology for both debris capture but also for orbital-traffic management. Here we will review important applications and potential end-use for tethers.

One of the simplest example use of tethers is to perform gravity-gradient maneuvers, where two satellites are attached by a tether and held vertically, where the lower satellite faces a slight different gravitational and centrifugal forces from the upper satellite [3]. The gravity gradient causes the bodies to be pulled apart and keeps the tether under tension and makes the system self-stabilized. Displacing the system from this equilibrium position can produce a restoring force where the two bodies snap from the tensile mode and start approaching each other.

A second application of a two-body system connected by tethers is through momentum exchange, where the upper-body gets boosted to higher elliptical orbit at the price of the lower body being lowered [4-5]. This approach involves the bodies being aligned to the gravity vector while being in orbit, followed by severing the tether to cause the upper body to raise itself into an elliptical orbit. Such an approach can be a low-cost, propellant-less method to boost payloads into higher orbits to avoid collision or for parking and storage.

A third major application is the use of multiple bodies attached by electro-conductive tethers to generate thrust by passing through the earth's magnetic field to produce a Lorentz force, a form of Electro-Motive Force (EMF) along the length of the tether [3, 11]. The induced EMF may amount to several kilovolts for an Electrodynamic Tether of 10-20 km. Depending on the direction of the tether and two bodies, the Lorentz force can act as a drag force to loweer the orbit of an object or be used as thrust to boost an object. The boosting occurs beyond Geostationary Orbit (GEO), where the electrodynamic tether lags behind the geomagnetic field and experience Lorentz force as a thrust. For all of these reason, there is an important advantage to grapple two free floating masses in space and attach tethers to them. In the following section, we will describe the tether dynamics and ways for using robots to attach tether between free flying objects in Earth orbit.

## 3. TETHER DYNAMICS MODELING

The tether connecting the robots can be most efficiently described as a flexible body as a series of point masses connected by massless springs and dampers in parallel as shown in Fig. 1. The tether geometry is represented by numbering the point masses as nodes and creating a graph $G = (N, E)$, where $N = \{1, 2, ..., n\}$ is a finite nonempty node set and $E \subset N \times N$ is an edge set of ordered pairs of nodes.

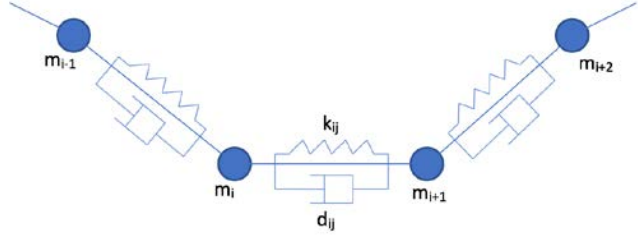

Fig. 1. Tether Model.

### 3.1 Flexible Dynamics Model

Using the Kelvin-Voigt model, the tether can be modeled as a viscoelastic material having the properties both of elasticity and viscosity through a combination of spring-dampers resulting in different tension laws. Tension on a rope element linking the $i^{th}$ node to the $j^{th}$ node can be expressed as Eq. (1).

$$T_{ij} = \begin{cases} [-k_{ij}(|r_{ij}| - l_0) - d_{ij}(v_{ij} \cdot \hat{r}_{ij})]\hat{r}_{ij} & if \ |r_{ij}| > l_0 \\ 0 & if \ |r_{ij}| \le l_0 \end{cases}$$

where, $k_{ij}$ is the stiffness parameter of the tether element $ij$ which depends on the material properties and geometry of the tether, $d_{ij}$ is the damping coefficient of the tether element $ij$. $r_{ij}$ and $v_{ij}$ are the relative position and velocity between the $i^{th}$ node and the $j^{th}$ node. $\hat{r}_{ij}$ is the normalized unit vector along the position vector. Also, $l_0$ is the nominal un-stretched length of the tether element. The stiffness parameter is directly proportional to the tether cross-sectional area $A$ and the Young's modulus $E$ and inversely proportional to the nominal length of the tether as shown in Eq. (2). Also, the damping coefficient depends on the damping ratio $\xi$, mass of the tether element between nodes $i$ and $j$ and the stiffness parameter as shown in Eq. (3).

$$k_{ij} = \frac{AE}{l_0}$$

$$d_{ij} = 2\xi \sqrt{m_{ij} k_{ij}}$$

## 3.2 Contact Dynamics Model

During the wrapping and docking phase, multiple contact events will occur between the tether and the target satellite and also among different part of the tethered system. As a result, efficient collision detection and accurate representation of contact dynamics becomes key to the fidelity of the simulation to reality. The target spacecraft is modeled as a convex polyhedra and the Gilbert, Johnson and Keerthi (GJK) collision detection algorithm is used to detect collision between the tether and the target satellite and also to calculate the penetration depth during every collision[12,13].

After detecting the collision, Hertz contact force model has been implemented to model the contact dynamics. When two bodies collide, local deformations occur resulting in penetration into each other's space. The penetration results in a pair of resistive contact forces acting on the two bodies in opposite directions. Every collision consists of a compression phase and a restitution phase which can be modeled as a non-linear spring-damper as shown in Eq. (4).

$$f_N = K\delta^n + d_c\dot{\delta}$$

where, $K$ is the stiffness parameter, which depends on the material properties and the local geometry of the contacting bodies, $\delta$ is the penetration depth, $d_c$ is the damping coefficient, $\dot{\delta}$ is the relative velocity of the contact points, projected on an axis normal to the contact surfaces and $n = 3/2$. For two colliding spheres with radii $R_i$ and $R_j$, the parameter $K$ can be determined as Eq. (5) and (6).

$$K = \frac{4}{3\pi(h_i + h_j)}\left(\frac{R_i R_j}{R_i + R_j}\right)^{\frac{1}{2}}$$

$$h_i = \frac{1 - v_k^2}{\pi E_k}; k = i, j$$

Where $v_k$ and $E_k$ are the Poisson's ratio and Young's modulus of each sphere. Also, the damping coefficient $d_c$ can be considered as a function of the penetration depth, $\delta$ and the hysteresis damping factor, $\mu$ as shown in Eq. (7) and (8).

$$d_c = \mu\delta^n$$

$$\mu = \frac{3K(1 - e^2)}{4^{(-)}\dot{\delta}}$$

where, $e$ is the coefficient of restitution and $^{(-)}\dot{\delta}$ is the penetration speed at the start of the compression phase.

## 3.3 Friction Model

Each collision between the tether and the target satellite results in a tangential frictional component of contact force which is computed using Coulomb's law of dry friction which opposes the relative motion. It has been experimentally found that the transition of friction force from zero to nonzero relative velocity is not instantaneous, but it takes place during a short period of time. This transition called the Stribeck effect is implemented to the equations of motion of the multibody system using the Anderson function to avoid stiction as shown in Eq. (9).

$$f_t = f_N\left(\mu_d + (\mu_s - \mu_d)e^{-\left(\frac{v_{i,j}}{v_s}\right)^p}\right)\tanh(k_t v_{i,j})$$

where, $\mu_s$ is the coefficient of static friction, $\mu_d$ is the coefficient of dynamic friction, $v_{ij} = v_i - v_j$ is the relative speed, $v_s$ is the coefficient of sliding speed that changes the shape of the decay in the Stribeck region, exponent $p$ affects the drop from static to dynamic friction and the parameter $k_t$ adjusts the slope of the curve from zero relative speed to the maximum static friction.

## 3.4 Aerodynamic Force Model

To compute the aerodynamic forces acting on the tether, the model presented by Aslanov and Ledkov is implemented[14]. One of the fundamental assumption of the model is that every half of the tether part connecting two point masses is considered rigid and hence moves at the same speed of the node. The aerodynamic force acting on a node $i$ can then be computed as shown in Eq. (10).

$$F_{ai} = \frac{\rho v_i d}{4} c_d \left(\frac{n_i}{r_{i,i-1}} + \frac{n_{i+1}}{r_{i+1,i}}\right)$$

where, $\rho$ is the atmospheric density, $v_i$ is the velocity of node $i$, $d$ is the tether diameter, $c_d$ is the drag coefficient, $r_{i,i-1}$ is the distance between node $i$ and $i - 1$, and $n_i = (v_i \times r_i) \times r_i$.

The block diagram to simulate the docking mechanism for the tethered system is shown in Fig. 2. The algorithm first computes the elastic and damping tension forces along with the aerodynamic forces acting on each node and then integrates the dynamic equations of motion to compute its positions and velocities.

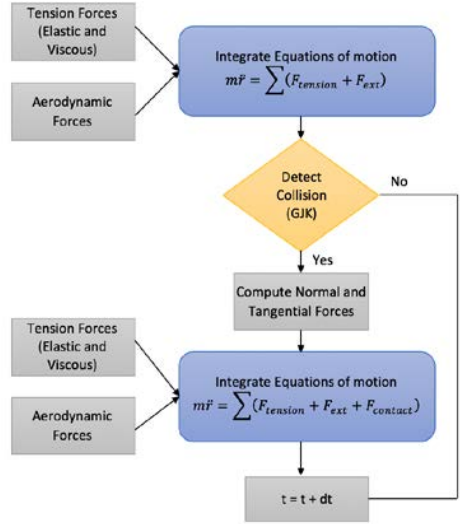

Fig. 2. Algorithm to solve Tether-Spacecraft interaction.

The collision detection algorithm is then carried out to detect impending collisions. The colliding nodes along with their penetration depth and relative velocities are computed and the corresponding contact normal and tangential forces are calculated which are then used to integrate the dynamic equations of motion.

## 4. CAPTURE AND DOCKING SIMULATION RESULTS

To fully analyze the dynamics of the tethered system and the contact model, simulations are performed on a simplified cubical target satellite (Fig. 3). The tethered system is modeled using 121 nodes, connected to four robots. The tethered system is deployed in a 'x' configuration with initial relative velocity w.r.t the target satellite of 15 m/s along the y-axis. The tether material is considered as Technora used to suspend the NASA Mars rover Opportunity from its parachute during descent.

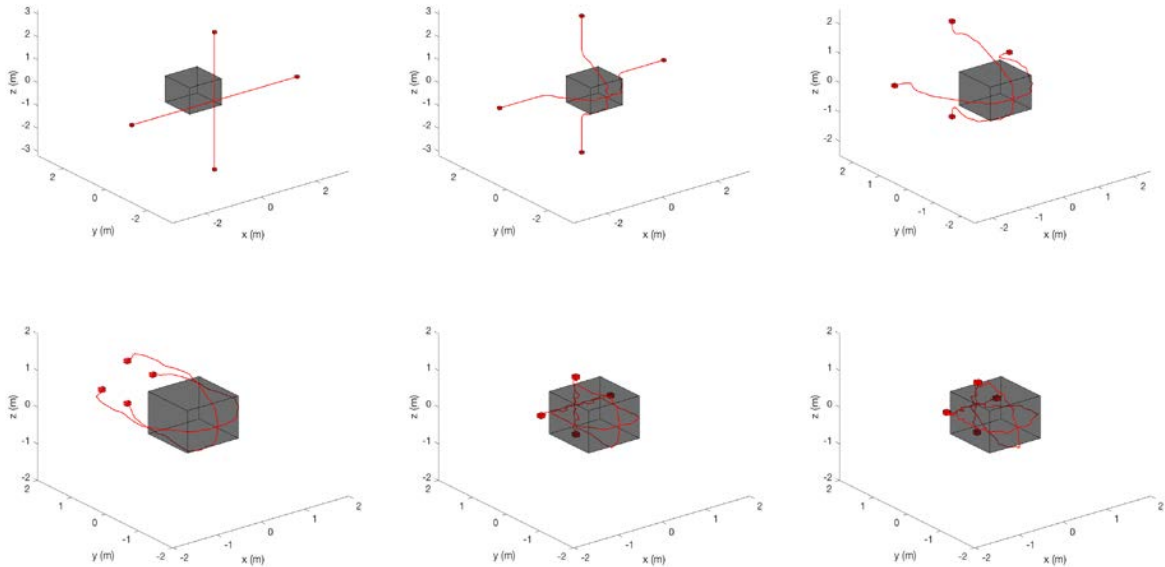

Fig. 3. Capture and Docking process at different time step with no rotation of target satellite.

For our simulation the Young's modulus of the tether is considered are 25 GPa, the damping ratio as 0.3 and the density as 1390 kg/m³. For the contact dynamics, the stiffness parameter is considered as 500 N/m and the damping coefficient as 0.5. For the friction model, the coefficient of static and dynamic friction are considered 0.7 and 0.5 respectively and the parameters as $v_s = 0.001$, $p = 2$, and $k_t = 10000$. The dimension of the target satellite is $1.15 \times 1.15 \times 1.15\ m$. Fig. 3 shows the capture and docking process at different time step. Further simulations were performed with the target satellite rotating with a constant angular velocity of [1 0.5 0.2] deg/s as shown in Fig. 4. It can be seen that the tethered robotic system was able to capture the target satellite.

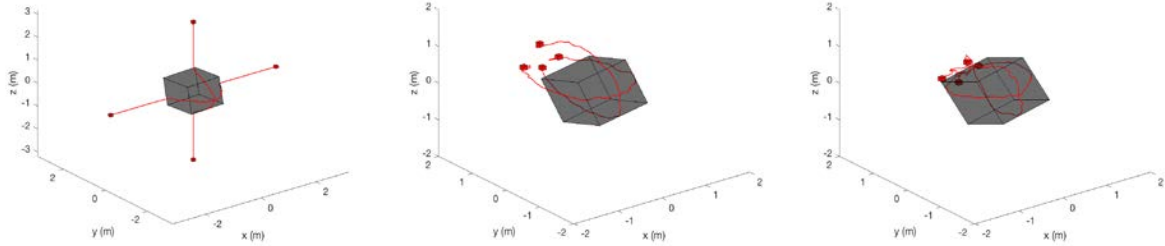

Fig. 4. Capture and Docking process at different time step with target satellite rotating with a constant angular velocity of [1, 0.5, 0.2] deg/s.

## 5. POST DOCKING OPERATIONS

There are many challenges to be addressed after the robots captures the satellite and docks onto it. Mitigating tangled tethers after docking is a major concern. Hence, the option exists for each robot to disconnect from its tether to avoid complex tangled scenarios. Furthermore, the tethers may be rolled up to shorten or lengthen the cord length between these robots. After docking with the target satellite, each robot secures itself on the satellites surface using spiny gripping actuators [2, 15]. The multi-robot system can form an 'x' configuration as shown in Fig. 5 (Left). It can then crawl onto the satellite's surface with each robot moving one by one using rolling (in conjunction with magnetic attraction) and hopping mobility capabilities as shown in Fig. 5 (Right) [2]. If any robot loses grip, the multirobot system with the remaining robots anchored to the surface of the target satellite keeps the entire system secure.

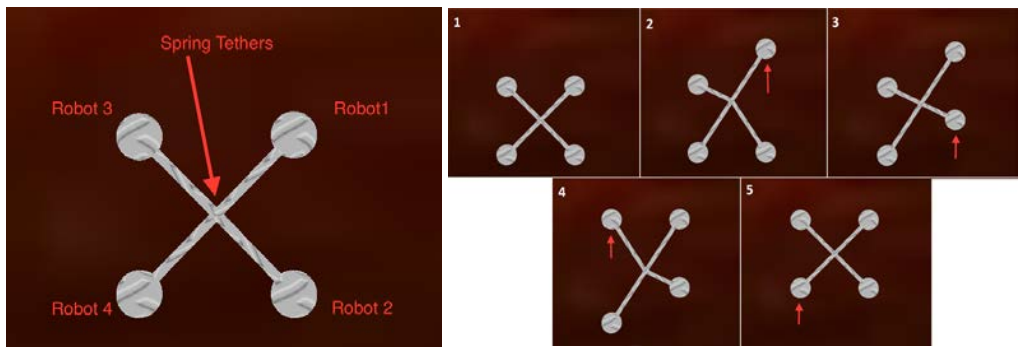

Fig. 5. (Left) 'x' configuration of multiple robots. (Right) Sequence of robot movement to crawl on the surface of the satellite.

The robots may need to autonomously traverse or explore a maze like environment on the surface of the target satellite, by following surface cues and incomplete maps of the target satellite [12]. This exploration would be critical to perform overall survey, including surveying for damage, identifying important resources onboard and for assisting in salvage planning. More immediate, the surveying effort would require the robots to identify a series of gripping points to move from one location to another. This is comparable to low-gravity climbing [13], however these robots encounter micro-gravity. If one of the robots were to slip, it would simply slide off the target satellite, however with the remaining robots firmly secure to the satellite, this bot will dangle from a tether and can then simply proceed to swing and grapple onto the surface of the satellite again.

This multirobot system can also be used to carry larger components and place them on a specific location on a target satellite. The rolling up of the tethers enables fine level position control of a larger object. Through this distributed controls approach, the risk is distributed, and a group of small robots can perform multiple servicing tasks on the satellite simultaneously.

## 6. CONCLUSION

In this paper, we present a new approach to satellite rendezvous, docking and repair using teams of tethered small robots. This approach is also readily applicable to salvage and dispose large space debris. Tethering a target satellite has some intrinsic advantages including the ability to perform momentum exchange to boost the orbit of a target satellite, use of Lorentz force to impose drag on the satellite to assist in de-orbiting or use of the tether to generate electricity to power the target satellite or repair activities. Importantly, using a tethered approach, a valuable servicing satellite can maintain a safe distance, while small dispensable microbots would capture, traverse and surveying the target satellite. Because these microbots will be of small size and mass, inadvertent impacts will minimize damage to the target satellite. Considering the promise of this approach to satellite servicing and debris capture/disposal, our efforts are now focused on designing prototype robots intended for this task, followed by high fidelity simulation of an end to end operations concept, followed by laboratory demonstration of critical tasks.